\documentclass[lettersize,journal]{IEEEtran}
\usepackage{amsmath,amsfonts}
\usepackage{algorithmic}
\usepackage{algorithm}
\usepackage{array}
\usepackage{textcomp}
\usepackage{stfloats}
\usepackage{url}
\usepackage{verbatim}
 \usepackage{longtable}
\usepackage{tabularx} 
\usepackage{multirow}
\usepackage[T1]{fontenc}
\usepackage{caption}
\usepackage{bbding} 
\usepackage[table]{xcolor}
\usepackage{xcolor}
\definecolor{rowgray}{gray}{0.93}
\usepackage{colortbl}
\usepackage{siunitx}
\usepackage{adjustbox}
\usepackage{tcolorbox}
\tcbuselibrary{skins, breakable}

\usepackage{booktabs, multirow, makecell, xspace, colortbl}
\usepackage{amsmath, amssymb}
\usepackage{amsthm}

\theoremstyle{remark}

\usepackage{pifont}

\usepackage{makecell}
\usepackage{graphicx}
\usepackage{subcaption}

\newcommand{\cmark}{\ding{51}}
\newcommand{\xmark}{\ding{55}}
\newcommand{\cmarkbold}{\textbf{\ding{51}}}
\usepackage{hyperref}
\hypersetup{hidelinks}
\newcolumntype{Y}{>{\raggedright\arraybackslash}X}
\newcolumntype{P}[1]{>{\raggedright\arraybackslash}p{#1}}

\begin{document}

\title{SpatialQuery: Benchmarking Geometry-Grounded Multi-Instance Spatial Reasoning in Vision-Language Models}

\newcommand{\ours}{\textsc{SpatialQuery}}
\newcommand{\method}{\ours}
\newcommand{\dataset}{\textsc{SpatialQuery-1M}}
\newcommand{\ourdataset}{\dataset}
\newcommand{\bev}{Bird's-Eye View}
\newcommand{\uacot}{UA-CoT}
\newcommand{\eg}{\textit{e.g.},}
\newcommand{\ie}{\textit{i.e.},}
\newcommand{\cf}{\textit{cf.}}
\newcommand{\etal}{\textit{et al.}}
\newcommand{\todo}[1]{\textcolor{red}{\textbf{[TODO: #1]}}}

\author{Hai Nguyen, Tung Vu, Cong Tran
\thanks{Hai Nguyen, Tung Vu, and Cong Tran are with the Posts and Telecommunications Institute of Technology, Hanoi, Vietnam.}
} 
\IEEEaftertitletext{\vspace{-2.5\baselineskip}}

\maketitle

\begin{abstract}
Vision-language models~(VLMs) achieve strong semantic understanding but
remain unreliable in metric spatial reasoning, particularly when queries
require comparing multiple instances of the same object category. We study
this problem through the \emph{Closest-Instance Distance Query}~(CIDQ), where
a model must identify the nearest visible candidate to a unique reference
object and estimate their gravity-aligned floor-plane distance. We introduce
\textbf{\ours{}}, a training-free framework for CIDQ reasoning from a single
RGB image, together with \textbf{\dataset{}}, a benchmark containing over one
million RGB-only question--answer pairs from 200 indoor scenes. \ours{}
recovers instance-level metric geometry and transforms it into a canonical
Bird's-Eye View through \emph{Scene Cubifying}, which represents objects as
uniformly sized, category-coded blocks to emphasize their relative
floor-plane locations. We further propose \emph{Uncertainty-Aware
Chain-of-Thought}~(\uacot{}) prompting, which incorporates geometry-derived
per-instance uncertainty into the VLM reasoning process. Without
task-specific fine-tuning or architectural modification, \ours{} with
Qwen3-VL-8B achieves a Floor-MAE of 0.259\,m, an Unc-Acc@0.3\,m of 90.5\%,
and a proximity-decision accuracy of 84.18\%, outperforming fine-tuned
spatial specialists, general-purpose VLMs, and closed-source frontier models.
Code, benchmark resources, and an interactive demo are available at
\href{https://namhai1810.github.io/SpatialQuery/}
{\url{https://namhai1810.github.io/SpatialQuery/}}.
\end{abstract}


\section{Introduction}
\label{sec:intro}

\IEEEPARstart{M}{ultimodal} generative models increasingly serve as interfaces
for perceiving, representing, and communicating information about physical
environments. Beyond recognizing scene content, their generated responses
must remain consistent with the underlying world geometry, particularly in
embodied AI, assistive systems, and interactive applications that require
metric-aware navigation or object interaction
~\cite{fan2022minedojo,yang2025guiding,song2025robospatial}.
However, despite strong semantic visual understanding, recent
Vision-Language Models~(VLMs)
~\cite{hurst2024gpt4o,bai2025qwen25vl,geminiteam2024gemini15}
remain unreliable at fine-grained metric reasoning
~\cite{chen2024spatialvlm,cheng2024spatialrgpt}.

Consider the question:
\emph{``What is the distance from the nearest chair to the TV?''}
Answering it requires more than estimating the distance between a predefined
object pair. A model must detect all visible chairs, recover their metric
locations relative to the reference object, compare the resulting distances,
and select the nearest instance. We formulate this setting as the
\textbf{Closest-Instance Distance Query (CIDQ)}, a multi-instance spatial
reasoning problem that jointly evaluates candidate-set aggregation,
closest-instance selection, and gravity-aligned floor-plane distance
estimation. Existing benchmarks predominantly focus on fixed object pairs and
therefore do not jointly assess these capabilities at scale.

CIDQ presents three main challenges. First, the final prediction depends on
recovering a complete and non-duplicated candidate set. Second, perspective
RGB images are poorly aligned with the horizontal geometry required for
indoor distance reasoning: camera tilt, perspective distortion, and object
elevation can make visually similar layouts correspond to different physical
distances. Third, monocular geometry is inherently uncertain, whereas most
spatial-reasoning systems treat estimated depth and object locations as
deterministic~\cite{bochkovskii2025depthpro,ma2024spatialpin}.

To address these challenges, we introduce \textbf{\ours{}}, a training-free
framework for geometry-grounded multi-instance reasoning from a single RGB
image. The framework first grounds candidate and reference objects and
recovers their metric floor-plane coordinates using monocular depth and
estimated camera geometry. It then constructs a compact world representation
through \emph{Scene Cubifying}, which renders detected instances as uniformly
sized, category-coded blocks on a canonical Bird's-Eye View~(BEV) canvas.
This abstraction suppresses appearance, scale, and perspective variations
while preserving the relative geometry required for CIDQ. Finally,
\emph{Uncertainty-Aware Chain-of-Thought}~(\uacot{}) prompting incorporates
geometry-derived per-instance uncertainty into the reasoning process, allowing
the generated prediction to reflect the reliability of the reconstructed
scene.

We further introduce \textbf{\dataset{}}, a large-scale benchmark containing
over one million RGB-only CIDQ question--answer pairs from 200 indoor scenes.
It supports two complementary tasks: closest-instance distance estimation and
proximity decision. Ground-truth camera parameters and 3D annotations are used
only for offline benchmark construction; evaluated models receive only an RGB
image and a natural-language query.

Our contributions are threefold:

\medskip
\noindent\textbf{(C1) Multi-instance metric spatial reasoning benchmark.}
We formulate CIDQ as a spatial reasoning problem over a variable-size set of
candidate instances and introduce \textbf{\dataset{}}, a benchmark containing
over one million RGB-only question--answer pairs with absolute
gravity-aligned floor-plane distance annotations.

\medskip
\noindent\textbf{(C2) Task-oriented top-down scene abstraction.}
We propose \emph{Scene Cubifying}, a gravity-aligned BEV interface
that renders monocularly reconstructed instance coordinates as
uniformly sized, category-coded blocks. The abstraction facilitates
multi-instance comparison by suppressing appearance, perspective,
and object-scale variations, while metric distances are computed
directly from the underlying floor-plane coordinates.

\medskip
\noindent\textbf{(C3) Geometry-aware multimodal reasoning.}
We introduce \uacot{}, a structured prompting strategy that incorporates
robust geometry-derived per-instance uncertainty into the VLM reasoning
process without task-specific fine-tuning or architectural modification.

\medskip
Under a strict zero-shot protocol, \ours{} with Qwen3-VL-8B achieves a
Floor-MAE of 0.259\,m, an Unc-Acc@0.3\,m of 90.5\%, and a
proximity-decision accuracy of 84.18\%, outperforming fine-tuned spatial
specialists, general-purpose VLMs, and closed-source frontier models.
Ablation results and the uncertainty-stratified analysis reported in
the supplementary material further demonstrate the complementary
effects of robust geometric refinement, canonical BEV representation,
and uncertainty-aware prompting.

\begin{table*}[t]
\centering
\caption{\textbf{Comparison of spatial reasoning methods.} ``Training-free'' means no fine-tuning or architectural change to the VLM backbone is required. ``Multi-inst. closest'' refers to the ability to reason over \textit{all} instances of a class to find the nearest one. ``Uncertainty'' indicates whether geometric uncertainty is propagated into the reasoning process. \cmark~= fully supported; $\sim$~= partially supported; \xmark~= not supported.}
\label{tab:comparison_methods}
\resizebox{\textwidth}{!}{%
\begin{tabular}{lccccccc}
\toprule
\textbf{Method} & \textbf{Depth Input} & \textbf{Training-free} & \textbf{Arch. unchanged} & \textbf{Multi-inst. closest} & \textbf{Uncertainty in reasoning} & \textbf{BEV visual input} & \textbf{Setting} \\
\midrule
SpatialVLM~\cite{chen2024spatialvlm} (CVPR'24)      & Estimated (auto)          & \xmark~fine-tune         & \xmark                 & \xmark & \xmark & \xmark & Indoor/Outdoor \\
SpatialRGPT~\cite{cheng2024spatialrgpt} (NeurIPS'24) & Relative depth map        & \xmark~fine-tune+plugin  & \xmark~depth connector & \xmark & \xmark & \xmark & Indoor/Outdoor/Sim \\
Ego3D-VLM (2025)~\cite{gholami2025spatialreasoningvisionlanguagemodels}                            & Metric depth (multi-view) & $\sim$~post-training     & \cmark                 & \xmark & $\sim$ & \xmark & Ego-centric multi-view \\
Talk2BEV~\cite{talk2bev} (2024)             & LiDAR (\textit{required}) & \cmark~pretrained VLM    & \cmark                 & \cmark & \xmark & \cmark & Autonomous driving \\
\midrule
\rowcolor{gray!20}
\textbf{Ours (proposed)} & \textbf{Metric depth (Depth Pro)} & \cmark~\textbf{plug-and-play} & \cmark & \cmark & \cmark & \cmark & \textbf{Single RGB indoor} \\
\bottomrule
\end{tabular}%
}
\end{table*}

\section{Related Work}
\label{sec:related}

\noindent\textbf{Spatial reasoning in VLMs and benchmarks.}
Despite strong semantic understanding, contemporary VLMs remain
unreliable in metric spatial reasoning
~\cite{fu2024blink,ramakrishnan2025spatial}.
SpatialVLM~\cite{chen2024spatialvlm},
SpatialRGPT~\cite{cheng2024spatialrgpt}, and
SpatialPIN~\cite{ma2024spatialpin} improve geometric grounding through
supervised fine-tuning, depth-aware architectural components, or
training-free geometric priors.
These approaches, however, predominantly reason over predefined object
pairs and do not jointly handle variable-cardinality candidate sets,
closest-instance selection, and geometry-derived uncertainty, which
are central to \textsc{cidq}.
Similarly, recent perspective-aware benchmarks mainly evaluate fixed
object configurations or relative spatial relations
~\cite{ma2025_3dsrbench,zhang2025do}.
\dataset{} extends this line of work to RGB-only, multi-instance
closest-object queries with absolute gravity-aligned floor-plane
distance annotations at scale.

\noindent\textbf{Perspective-normalized spatial representations.}
Egocentric views introduce viewpoint bias and hinder allocentric
reasoning~\cite{goral2024seeing}. APC~\cite{lee2025apc} transforms
coarse 3D abstractions into a selected viewer's egocentric frame,
while Talk2BEV~\cite{talk2bev} enriches driving-scene BEV maps with
semantic and language cues. In contrast, our \emph{Scene Cubifying}
maps monocularly reconstructed instances into a shared,
gravity-aligned floor-plane frame and renders them as uniformly sized,
category-coded cuboids. The BEV serves only as a compact visual
interface, whereas metric distances are computed directly from the
retained floor-plane coordinates.

\noindent\textbf{Visual prompting and uncertainty-aware reasoning.}
Chain-of-thought and visual prompting methods improve reasoning by
externalizing intermediate evidence
~\cite{wei2022chain,yang2023setofmarks,hu2024visualsketchpad}.
However, these formulations generally assume that the visual or
geometric evidence supplied to the reasoning model is deterministic
and equally reliable.
Our \uacot{} instead conditions reasoning on both the reconstructed BEV
layout and per-instance dispersion $\sigma_i^{\mathrm{BEV}}$, allowing
closest-instance selection and distance prediction to reflect the
reliability of monocular scene reconstruction. 

\section{Task Definition}
\label{sec:task}

A \textbf{Closest-Instance Distance Query}~(CIDQ) evaluates whether
a model can reason over multiple visible instances of the same
semantic category and identify the one nearest to a designated
reference object. Unlike conventional pairwise spatial queries,
CIDQ requires candidate-set aggregation, closest-instance selection,
and metric distance estimation under a unified formulation.
We consider static, single-floor indoor scenes and provide only a
single RGB image and a natural-language query as model inputs; no
depth, camera parameters, or multi-view observations are available
at inference time.

\subsection{CIDQ Formulation}
\label{sec:task:formal}

A CIDQ configuration is specified by
\[
  q=(I,C_q,o_r),
  \qquad
  C_r:=\mathrm{label}(o_r),
  \qquad
  C_q\neq C_r,
\]
where $I\in\mathbb{R}^{H\times W\times3}$ is an RGB image,
$C_q$ is the candidate category, and $o_r$ is the unique visible
reference instance of category $C_r$. The tuple $q$ encodes the
structured task semantics; the evaluated model receives only $I$ and
a natural-language realization $Q$ of these semantics.

The qualified visible instance sets satisfy
\[
  \mathcal{O}_{C_r}(I)=\{o_r\},
  \qquad
  \mathcal{O}_{C_q}(I)=\{o_i\}_{i=1}^{n},
  \qquad n\geq2.
\]
The benchmark-specific visibility and validity criteria are described
in Section~\ref{sec:dataset}. The unique reference and multiple
candidates distinguish CIDQ from conventional fixed-pair estimation.

Let $\mathbf{c}_i,\mathbf{c}_r\in\mathbb{R}^3$ denote the
gravity-aligned centroids of candidate $o_i$ and reference $o_r$,
respectively. For $\mathbf{c}=(X,Y,Z)^\top$, with the $Y$-axis aligned
with gravity, define the floor-plane projection
\[
  \Pi_{\mathrm{floor}}(\mathbf{c})=(X,Z)^\top.
\]

The ground-truth closest-candidate index and projected-centroid
floor-plane distance are
\begin{align}
  i_q^\star
  &=
  \operatorname*{arg\,min}_{i\in\{1,\ldots,n\}}
  \left\|
    \Pi_{\mathrm{floor}}(\mathbf{c}_i)
    -
    \Pi_{\mathrm{floor}}(\mathbf{c}_r)
  \right\|_2,
  \label{eq:cidq_inst_index}
  \\
  d_q^\star
  &=
  \left\|
    \Pi_{\mathrm{floor}}(\mathbf{c}_{i_q^\star})
    -
    \Pi_{\mathrm{floor}}(\mathbf{c}_r)
  \right\|_2,
  \label{eq:cidq_dist}
\end{align}
where $d_q^\star\in\mathbb{R}_{\geq0}$ is expressed in metres and
$o_{i_q^\star}$ is the corresponding closest instance.

CIDQ supports two complementary tasks.

\smallskip
\noindent\textbf{T1: Closest-Instance Distance Estimation.}\quad
Given $I$ and a natural-language query $Q$ realizing $q$, predict the
metric floor-plane distance $\hat d_q$ corresponding to $d_q^\star$.

\smallskip
\noindent\textbf{T2: Proximity Decision.}\quad
For a threshold $\tau>0$, define the threshold-augmented query
\[
  q_\tau=(q,\tau)=(I,C_q,o_r,\tau).
\]
Given $I$ and a natural-language query $Q$ realizing $q_\tau$, predict
$\hat y_{q_\tau}\in\{0,1\}$ for
\begin{equation}
  y_{q_\tau}^\star
  =
  \mathbf{1}\!\left[d_q^\star\leq\tau\right].
  \label{eq:cidq_decision}
\end{equation}

\subsection{Evaluation Metrics}
\label{sec:task:metrics}

For each method and metric,
$\mathcal{Q}_{\mathrm{eval}}^{\mathrm{T1}}$ and
$\mathcal{Q}_{\mathrm{eval}}^{\mathrm{T2}}$ denote evaluation sets for T1 and T2, respectively. Their construction and sizes are specified in
Section~\ref{sec:exp:setup}.

\noindent\textbf{Distance-estimation metrics.}\quad
For T1, Floor-MAE is computed over the available original-image
distance predictions:
\begin{equation}
  \mathrm{Floor\text{-}MAE}
  =
  \frac{1}{
    |\mathcal{Q}_{\mathrm{eval}}^{\mathrm{T1}}|
  }
  \sum_{q\in\mathcal{Q}_{\mathrm{eval}}^{\mathrm{T1}}}
  \left|
    \hat d_q-d_q^\star
  \right|,
  \label{eq:mae}
\end{equation}
We additionally report \textbf{Acc@$\delta$}, the fraction of predictions satisfying $\left|\hat d_q-d_q^\star\right|\leq\delta_q$, using $\delta_q=0.1d_q^\star$ for Acc@10\% and $\delta_q=0.2\,\mathrm{m}$ for Acc@0.2\,m; the corresponding Unc-Acc metrics use the same base tolerances.

\smallskip
\noindent\textbf{Uncertainty-conditioned accuracy.}\quad
When a method reports a non-negative distance uncertainty
$\hat u_q$, we additionally compute
\begin{equation}
  \mathrm{Unc\text{-}Acc@}\delta
  =
  \frac{1}{
    |\mathcal{Q}_{\mathrm{eval}}^{\mathrm{T1}}|
  }
  \sum_{q\in\mathcal{Q}_{\mathrm{eval}}^{\mathrm{T1}}}
  \mathbf{1}\!\left[
    \left|\hat d_q-d_q^\star\right|
    \leq
    \delta_q+\hat u_q
  \right].
  \label{eq:unc_acc}
\end{equation}
This auxiliary metric measures error coverage after conditioning the
acceptance tolerance on the uncertainty reported by each method.
It is not interpreted as a statistically calibrated confidence
interval. For \ours{}, $\hat u_q$ is instantiated by the
geometry-derived query uncertainty $\sigma_q$ defined in
Section~\ref{sec:phase5}.

\smallskip
\noindent\textbf{Proximity-decision accuracy.}\quad
For T2, we report standard binary accuracy, defined as the proportion of evaluation queries for which the predicted proximity label $\hat y_{q_\tau}$ matches the corresponding ground-truth label $y_{q_\tau}^\star$ over $\mathcal{Q}_{\mathrm{eval}}^{\mathrm{T2}}$.

The formulation above defines the CIDQ prediction targets and evaluation criteria independently of any particular inference pipeline. Section~\ref{sec:method} describes how \ours{} estimates $\hat d_q$, $\sigma_q$, and $\hat y_{q_\tau}$ from an RGB--query pair, while Section~\ref{sec:dataset} explains how the corresponding queries and ground-truth labels are constructed at scale.

\section{The \ours{} Framework}
\label{sec:method}
\subsection{Overview}
\label{sec:method:overview}

Given a single RGB image $I$ and a natural-language query $Q$,
\ours{} parses the candidate category $C_q$, reference category $C_r$,
and, for T2, threshold $\tau$. It then grounds the relevant instances,
reconstructs their metric floor-plane geometry, and encodes the scene
as a canonical gravity-aligned Bird's-Eye View~(BEV).

\begin{figure*}[!ht]
  \centering
  \includegraphics[width=\textwidth]{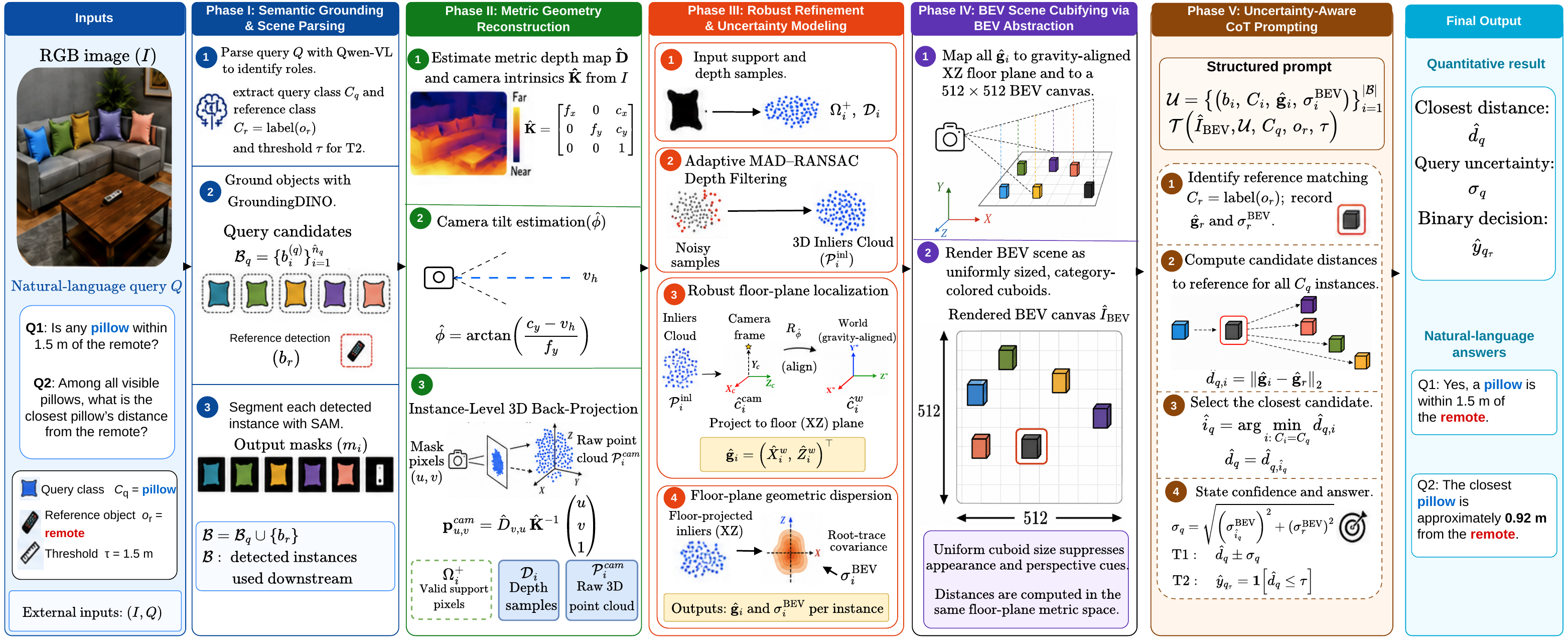}
  \caption{%
    \textbf{\ours{} Pipeline.}
    \emph{Phase~I}: Qwen-VL~\cite{bai2023qwen} identifies entity
    roles; GroundingDINO~\cite{liu2024grounding} and
    SAM~\cite{carion2025sam3segmentconcepts} localize and segment all instances.
    \emph{Phase~II}: Depth Pro~\cite{bochkovskii2025depthpro} estimates
zero-shot metric depth and camera intrinsics; a geometric estimator
recovers the camera pitch angle~$\hat{\phi}$; instance pixels are
back-projected into metric camera-frame point clouds.
    \emph{Phase~III}: MAD-based RANSAC filters unreliable depth support;
the refined geometry is pitch-corrected to obtain gravity-aligned
floor-plane coordinates and uncertainty $\sigma_i^{\mathrm{BEV}}$.
    \emph{Phase~IV}: Tilt-corrected 3D positions are projected onto a
    canonical top-down BEV canvas; each object is rendered as a
    color-coded cuboid (\emph{Scene Cubifying}).
    \emph{Phase~V}: The BEV abstraction and per-instance BEV uncertainty~$\sigma_i^{\mathrm{BEV}}$ are injected into a structured \uacot{}
    prompt, guiding the VLM toward a geometry-aware final answer.}
  \label{fig:pipeline}
\end{figure*}

As illustrated in Fig.~\ref{fig:pipeline}, the five-phase pipeline
combines the BEV representation with geometry-derived uncertainty to
predict the closest-instance distance $\hat d_q$, query uncertainty
$\sigma_q$, and, for T2, proximity decision $\hat y_{q_\tau}$, without
task-specific fine-tuning or architectural modification.

\subsection{Phase I: Semantic Grounding and Scene Parsing}
\label{sec:phase1}

Qwen-VL~\cite{bai2023qwen} parses $Q$ into the candidate category
$C_q$, reference category $C_r$, and, when applicable, threshold
$\tau$. GroundingDINO~\cite{liu2024grounding} then localizes all
visible $C_q$ candidates and the object matching $C_r$. Because CIDQ
assumes a unique visible reference, duplicate reference detections are
removed by class-wise non-maximum suppression, and the
highest-confidence detection is retained as $b_r$. Thus, $b_r$ is the
inference-time grounding of the task-level reference instance $o_r$.
SAM~\cite{carion2025sam3segmentconcepts} converts each retained box
into an instance mask.

Let
$\mathcal{B}_q=\{b_i^{(q)}\}_{i=1}^{\hat n_q}$
denote the retained query-class boxes, where
$\hat n_q:=|\mathcal{B}_q|$ and $b_i^{(q)}$ is the
$i$-th candidate box. The complete detected-instance set is
\begin{equation}
\begin{aligned}
  \mathcal{B}
  &=
  \mathcal{B}_q\cup\{b_r\},
  \qquad
  |\mathcal{B}|=\hat n_q+1,
  \\
  \mathcal{S}_0
  &=
  \left\{
    (b_i,C_i,m_i)
  \right\}_{i=1}^{|\mathcal{B}|},
  \qquad
  m_i
  \subseteq
  \{1,\ldots,W\}\times\{1,\ldots,H\}.
\end{aligned}
\label{eq:full_detected_instances}
\end{equation}
Here, $C_i$ and $m_i$ denote the semantic category and instance mask
associated with $b_i$, respectively, and $r$ is the unique index
assigned to the reference detection $b_r$. The masks provide
instance-specific pixel support for metric reconstruction in
Phase~II.

\subsection{Phase II: Metric Geometry Reconstruction}
\label{sec:phase2}

Given the RGB image $I$, Depth Pro~\cite{bochkovskii2025depthpro}
predicts a dense metric-depth map
$\hat{\mathbf D}\in\mathbb{R}_{\geq 0}^{H\times W}$
and the camera intrinsic matrix
\begin{equation}
  \hat{\mathbf K}
  =
  \begin{bmatrix}
    f_x & 0   & c_x \\
    0   & f_y & c_y \\
    0   & 0   & 1
  \end{bmatrix},
  \label{eq:estimated_intrinsics}
\end{equation}
where $f_x$ and $f_y$ are the horizontal and vertical focal lengths,
and $(c_x,c_y)$ is the principal point. We index an image pixel by
$(u,v)$, where $u\in\{1,\ldots,W\}$ increases rightward and
$v\in\{1,\ldots,H\}$ increases downward. Accordingly,
$\hat D_{v,u}$ denotes the predicted metric depth at pixel $(u,v)$.

We adopt a right-handed camera coordinate frame in which
$X^{\mathrm{cam}}$ points rightward, $Y^{\mathrm{cam}}$ points
downward, and $Z^{\mathrm{cam}}$ points forward along the optical axis.
Because the input camera may not be level, a depth-derived geometric
estimator first recovers the horizon row $v_h$. Under the downward
image-row convention, the downward-positive camera pitch is
\begin{equation}
  \hat{\phi}
  =
  \arctan\!\left(
    \frac{c_y-v_h}{f_y}
  \right),
  \label{eq:tilt}
\end{equation}
where $\hat{\phi}=0$ corresponds to a level camera and
$\hat{\phi}>0$ denotes a downward-looking viewpoint. Additional
coordinate conventions, pitch-correction details, and the complete
back-projection formulation are provided in the supplementary material.

For each retained detection
$i\in\{1,\ldots,|\mathcal{B}|\}$, let $m_i$ denote its instance
mask obtained in Phase~I. We first define its valid depth support as
\begin{equation}
  \Omega_i^{+}
  =
  \left\{
    (u,v)\in m_i:
    \hat D_{v,u}>0
  \right\},
  \label{eq:valid_depth_support}
\end{equation}
where $\Omega_i^{+}$ contains all mask pixels for which a positive
metric-depth prediction is available. Each valid pixel is then
back-projected into the camera frame as
\begin{equation}
  \mathbf p_{u,v}^{\mathrm{cam}}
  =
  \hat D_{v,u}\,
  \hat{\mathbf K}^{-1}
  \begin{bmatrix}
    u & v & 1
  \end{bmatrix}^{\!\top}
  \in\mathbb{R}^{3},
  \qquad
  (u,v)\in\Omega_i^{+}.
  \label{eq:pixel_backprojection}
\end{equation}

The scalar depth samples and the corresponding raw camera-frame point
cloud of instance $i$ are collected as
\begin{equation}
\begin{aligned}
  \mathcal D_i
  &:=
  \left(
    \hat D_{v,u}
  \right)_{(u,v)\in\Omega_i^{+}},
  \\
  \mathcal P_i^{\mathrm{cam}}
  &:=
  \left\{
    \mathbf p_{u,v}^{\mathrm{cam}}:
    (u,v)\in\Omega_i^{+}
  \right\}
  \subset\mathbb{R}^{3}.
\end{aligned}
\label{eq:raw_instance_geometry}
\end{equation}
Here, $\mathcal D_i$ is the collection of instance-specific depth
samples, while $\mathcal P_i^{\mathrm{cam}}$ is their corresponding
metric 3D support in the camera coordinate frame.

Phase~II therefore transforms each image-space instance mask $m_i$
into a raw metric point cloud $\mathcal P_i^{\mathrm{cam}}$, together
with its depth samples $\mathcal D_i$ and the estimated camera pitch
$\hat{\phi}$. Phase~III subsequently removes unreliable depth support,
estimates a robust instance location, and rotates the retained geometry
into a gravity-aligned metric frame.

\subsection{Phase III: Robust Refinement and Uncertainty Modeling}
\label{sec:phase3}

For each detected instance $i$, Phase~II provides the valid depth
support $\Omega_i^{+}$, depth samples $\mathcal{D}_i$, and raw
camera-frame point cloud $\mathcal{P}_i^{\mathrm{cam}}$. Because
instance masks may still include depth spikes, boundary leakage, and
background pixels, we refine the depth support before estimating the
object location.

\noindent\textbf{Adaptive depth scale.}\quad
We estimate the instance depth center and dispersion using the median
and median absolute deviation~(MAD):
\begin{equation}
\begin{aligned}
  \mu_i^{\hat D}
  &=
  \operatorname{median}
  \left(
    \mathcal{D}_i
  \right),
  \\
  s_i^{\hat D}
  &=
  1.4826\,
  \operatorname{median}_{d\in\mathcal{D}_i}
  \left|
    d-\mu_i^{\hat D}
  \right|
  +
  \varepsilon_{\mathrm{num}}.
\end{aligned}
\label{eq:mad_depth_scale}
\end{equation}
Here, $s_i^{\hat D}$ defines an instance-adaptive depth scale, and
$\varepsilon_{\mathrm{num}}>0$ prevents a degenerate threshold for
nearly constant depth support.

\noindent\textbf{Dominant depth-mode selection.}\quad
To separate the object surface from competing background depths, we
perform one-dimensional RANSAC. At iteration $k$, a hypothesis
$z_i^{(k)}$ is sampled from $\mathcal{D}_i$, with MAD-scaled consensus
\begin{equation}
\begin{aligned}
  \Omega_i^{(k)}
  &=
  \left\{
    (u,v)\in\Omega_i^{+}:
    \left|
      \hat D_{v,u}-z_i^{(k)}
    \right|
    \leq
    \lambda_{\mathrm{MAD}}s_i^{\hat D}
  \right\},
  \\
  k_i^\star
  &=
  \operatorname*{arg\,max}_{1\leq k\leq K_{\mathrm{iter}}}
  \left|
    \Omega_i^{(k)}
  \right|,
  \qquad
  \Omega_i^{\mathrm R}
  =
  \Omega_i^{(k_i^\star)}.
\end{aligned}
\label{eq:mad_ransac_consensus}
\end{equation}
Here, $\lambda_{\mathrm{MAD}}>0$ controls the acceptance interval,
$K_{\mathrm{iter}}$ is the number of hypotheses, and
$\Omega_i^{\mathrm R}$ is the maximum consensus representing the
dominant visible depth mode.

Because the selected hypothesis may not coincide with the mode center,
we re-estimate it from the consensus median and apply a second
filtering pass:
\begin{equation}
\begin{aligned}
  \bar{\mu}_i^{\hat D}
  &=
  \operatorname{median}_{(u,v)\in\Omega_i^{\mathrm R}}
  \hat D_{v,u},
  \\
  \Omega_i^{\mathrm{inl}}
  &=
  \left\{
    (u,v)\in\Omega_i^{\mathrm R}:
    \left|
      \hat D_{v,u}-\bar{\mu}_i^{\hat D}
    \right|
    \leq
    \lambda_{\mathrm{MAD}}s_i^{\hat D}
  \right\},
  \\
  \mathcal{P}_i^{\mathrm{inl}}
  &=
  \left\{
    \mathbf p_{u,v}^{\mathrm{cam}}:
    (u,v)\in\Omega_i^{\mathrm{inl}}
  \right\}.
\end{aligned}
\label{eq:inlier_point_cloud}
\end{equation}
The first pass selects the dominant depth mode, while the second trims
its residual outliers. The resulting
$\mathcal{P}_i^{\mathrm{inl}}$ is used for object localization.

\noindent\textbf{Robust floor-plane localization.}\quad
We estimate the camera-frame object location using the coordinate-wise
median of the filtered point cloud, then correct the camera pitch and
project the result onto the floor plane:
\begin{equation}
\begin{aligned}
  \hat{\mathbf c}_i^{\mathrm{cam}}
  &=
  \operatorname{median}_{\mathrm{coord}}
  \left(
    \mathcal{P}_i^{\mathrm{inl}}
  \right),
  \\
  \hat{\mathbf c}_i^{\mathrm w}
  &=
  \mathbf R_{\hat\phi}
  \hat{\mathbf c}_i^{\mathrm{cam}}
  =
  \left(
    \hat X_i^{\mathrm w},
    \hat Y_i^{\mathrm w},
    \hat Z_i^{\mathrm w}
  \right)^\top,
  \\
  \hat{\mathbf g}_i
  &=
  \Pi_{\mathrm{floor}}
  \left(
    \hat{\mathbf c}_i^{\mathrm w}
  \right)
  =
  \left(
    \hat X_i^{\mathrm w},
    \hat Z_i^{\mathrm w}
  \right)^\top
  \in\mathbb{R}^{2}.
\end{aligned}
\label{eq:robust_instance_location}
\end{equation}
Here, $\operatorname{median}_{\mathrm{coord}}$ operates independently
on the three coordinates, $\mathbf R_{\hat\phi}$ corrects the camera
pitch, and $\hat{\mathbf g}_i$ is the gravity-aligned metric
floor-plane coordinate.

\noindent\textbf{Floor-plane geometric dispersion.}\quad
We measure geometric reliability in the same floor-plane coordinate
system used by CIDQ. The retained points are rotated and projected as
\begin{equation}
\begin{aligned}
  \mathcal{G}_i^{\mathrm{inl}}
  &=
  \left\{
    \Pi_{\mathrm{floor}}
    \left(
      \mathbf R_{\hat\phi}\mathbf p
    \right):
    \mathbf p\in\mathcal{P}_i^{\mathrm{inl}}
  \right\}
  \subset\mathbb{R}^{2},
  \\
  \sigma_i^{\mathrm{BEV}}
  &=
  \sqrt{
    \operatorname{tr}
    \left[
      \operatorname{Cov}
      \left(
        \mathcal{G}_i^{\mathrm{inl}}
      \right)
    \right]
  }.
\end{aligned}
\label{eq:bev_uncertainty}
\end{equation}
The covariance trace aggregates dispersion along both floor-plane
axes, while its square root expresses the score in metric units.
Thus, $\sigma_i^{\mathrm{BEV}}$ serves as an instance-level geometric
reliability proxy rather than a calibrated probabilistic confidence
interval.

Phase~III returns
$(\hat{\mathbf g}_i,\sigma_i^{\mathrm{BEV}})$ for each detected
instance. These outputs are used for BEV construction in Phase~IV and
uncertainty-aware reasoning in Phase~V. Exact hypothesis sampling,
parameter settings, and degenerate-case handling are provided in the
supplementary material.

\subsection{Phase IV: Scene Cubifying via BEV Abstraction}
\label{sec:phase4}

The refined floor-plane coordinates constitute the metric representation
used for CIDQ distance computation. To provide the VLM with a compact
visual interface, we additionally render these coordinates on a
$512\times512$ gravity-aligned top-down canvas. All instances are mapped
using the same scene-level normalization and represented as uniformly
sized, category-coded blocks, while the reference instance is highlighted
by a contrasting border.

This \emph{Scene Cubifying} abstraction suppresses RGB appearance,
perspective-dependent scale, and object-shape cues while retaining the
relative candidate--reference layout. The resulting image
$\hat I_{\mathrm{BEV}}$ is used only as visual context: pixel distances
are never interpreted metrically, and all numerical distances and
uncertainty values are computed from the underlying coordinates
$\hat{\mathbf g}_i$ and dispersion scores
$\sigma_i^{\mathrm{BEV}}$. Phase~V receives
$\hat I_{\mathrm{BEV}}$ together with this structured instance geometry.

\subsection{Phase V: Uncertainty-Aware Chain-of-Thought
Prompting}
\label{sec:phase5}

Using these outputs, \uacot{} supplies the VLM with both the canonical
BEV abstraction and the geometry associated with every localized
instance:
\begin{equation}
  \mathcal{U}
  =
  \left\{
    \left(
      b_i,\,
      C_i,\,
      \hat{\mathbf g}_i,\,
      \sigma_i^{\mathrm{BEV}}
    \right)
  \right\}_{i=1}^{|\mathcal B|}.
  \label{eq:prompt_instance_set}
\end{equation}
The structured prompt
$\mathcal{T}(\hat I_{\mathrm{BEV}},\mathcal U,C_q,o_r,\tau)$
identifies $b_r$ as the inference-time grounding of the task-level
reference $o_r$. The original RGB image is not supplied during this
reasoning phase, forcing candidate comparison to operate on the
canonical floor-plane representation. For T1, the threshold argument
$\tau$ is omitted.

The complete semantic-role extraction and \uacot{} inference prompts
are provided in the supplementary material.

For the candidate-index set
$\mathcal I_q:=\{i\in\{1,\ldots,|\mathcal{B}|\}:C_i=C_q\}$,
the prompt computes all candidate--reference distances and selects
the minimum:
\begin{equation}
\begin{aligned}
  \hat d_{q,i}
  &=
  \left\|
    \hat{\mathbf g}_i
    -
    \hat{\mathbf g}_r
  \right\|_2,
  \qquad i\in\mathcal I_q,
  \\
  \hat i_q
  &=
  \operatorname*{arg\,min}_{i\in\mathcal I_q}
  \hat d_{q,i},
  \qquad
  \hat d_q
  =
  \hat d_{q,\hat i_q}.
\end{aligned}
\label{eq:predicted_closest}
\end{equation}
Here, $\hat i_q$ is the predicted closest-candidate index and
$\hat d_q$ is its predicted gravity-aligned floor-plane distance.

After candidate selection, the dispersion scores of the selected
candidate and reference are aggregated by root-sum-square:
\begin{equation}
  \sigma_q
  =
  \sqrt{
    \left(
      \sigma_{\hat i_q}^{\mathrm{BEV}}
    \right)^2
    +
    \left(
      \sigma_r^{\mathrm{BEV}}
    \right)^2
  }.
  \label{eq:query_uncertainty}
\end{equation}
The T1 response is reported as
$\hat d_q\pm\sigma_q$ metres, and the uncertainty supplied to
Eq.~\eqref{eq:unc_acc} is instantiated as
$\hat u_q=\sigma_q$. This retains the point prediction
$\hat d_q$ for conventional distance metrics while exposing the
reliability of the reconstructed candidate--reference geometry.

For T2, the same closest-instance estimate produces the binary
decision
\begin{equation}
  \hat y_{q_\tau}
  =
  \mathbf{1}\!\left[
    \hat d_q\leq\tau
  \right].
  \label{eq:predicted_binary}
\end{equation}
Here, $\hat y_{q_\tau}=1$ denotes \textsc{Yes}, whereas
$\hat y_{q_\tau}=0$ denotes \textsc{No}. This completes the inference pathway from an RGB-query pair to
$\hat d_q$, $\sigma_q$, and, for T2, $\hat y_{q_\tau}$.

To evaluate these predictions against independently constructed
ground truth, we next build a large-scale CIDQ benchmark from offline
3D annotations.

\begin{figure*}[t]
  \centering
  \includegraphics[width=\textwidth]{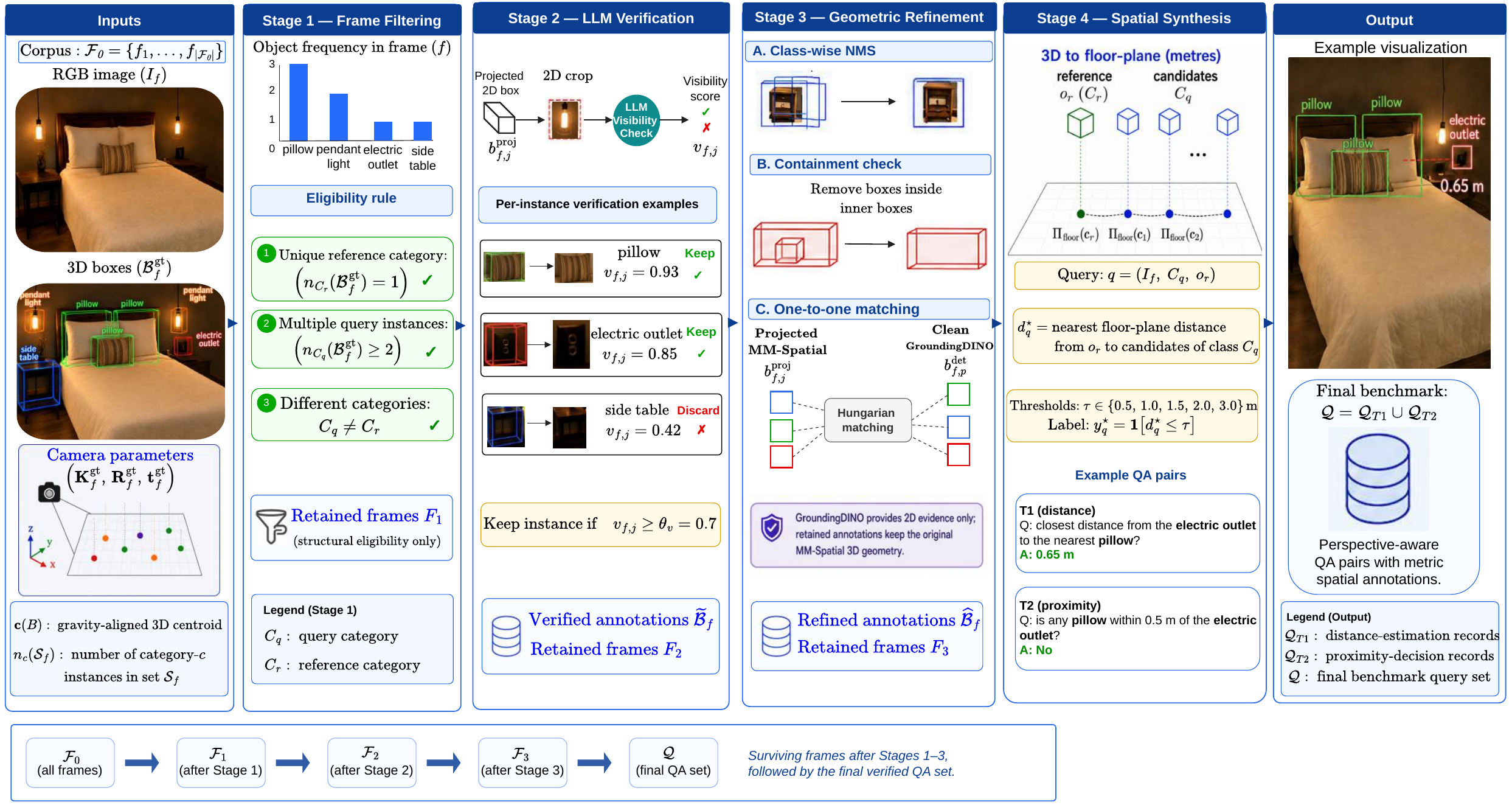}
  \caption{%
    \textbf{Four-stage QA synthesis pipeline for \dataset{}.}
    Indoor frames from MM-Spatial pass through:
    \textbf{Stage~1} (\emph{Frame Filtering}) --- scenes are
    retained based on object frequency analysis, requiring at
    least one unique reference instance and at least one
    category with at least two co-occurring instances;
    \textbf{Stage~2} (\emph{LLM Verification}) --- ChatGPT
    zero-shot prompting assigns probabilistic visibility scores
    to each instance, discarding occluded or semantically noisy
    detections;
    \textbf{Stage~3} (\emph{Geometric Refinement}) --- GroundingDINO
    re-detections are deduplicated via IoU thresholding and a
    box-in-box containment check, improving annotation consistency;
    \textbf{Stage~4} (\emph{Spatial Synthesis}) --- gravity-aligned
3D centroids are projected onto the floor plane, and Euclidean
distances between their BEV coordinates yield absolute floor-plane
metric ground truth, from which CIDQ question--answer pairs are
generated.
  }
  \label{fig:dataset_pipeline}
\end{figure*}

\section{The \dataset{} Benchmark}
\label{sec:dataset}

\textbf{\dataset{}} instantiates CIDQ at scale using independently
curated ground-truth geometry. Existing spatial-reasoning benchmarks
do not simultaneously provide RGB-only model inputs, multi-instance
closest-object queries, and absolute gravity-aligned metric ground
truth (Table~\ref{tab:dataset_comparison}).

To address this gap, \dataset{} contains over one million indoor CIDQ
question--answer pairs. As illustrated in
Fig.~\ref{fig:dataset_pipeline}, its four-stage construction pipeline
curates visually valid object annotations and maps them to the
task-level queries and ground-truth labels defined in
Section~\ref{sec:task}. Ground-truth camera parameters and 3D
annotations are used only for offline benchmark construction and are
never exposed to the evaluated models.

\subsection{Source Data and Construction}
\label{sec:dataset:source}

\dataset{} is constructed from the indoor subset of
MM-Spatial~\cite{daxberger2025mmspatial}, which provides
instance-level 3D annotations and calibrated camera geometry for
scenes derived from CA-1M~\cite{lazarow2025cubifyanything}.
CA-1M contains more than 1{,}000 indoor scenes and 439K annotated
objects across common room types, including bedrooms, living rooms,
kitchens, offices, corridors, and dining rooms.

Let $\mathcal{F}_0$ denote the source frames before eligibility
filtering. Each frame $f\in\mathcal{F}_0$ provides an RGB image
$I_f$, an instance-level annotation set
$\mathcal{B}^{\mathrm{gt}}_f$, and calibrated camera parameters
$(\mathbf{K}^{\mathrm{gt}}_f,
\mathbf{R}^{\mathrm{gt}}_f,
\mathbf{t}^{\mathrm{gt}}_f)$.
Each annotation $B\in\mathcal{B}^{\mathrm{gt}}_f$ has a semantic
label $\mathrm{label}(B)$ and a gravity-aligned 3D centroid
$\mathbf{c}(B)\in\mathbb{R}^3$.
For an annotation set $\mathcal S_f$, let $n_c(\mathcal S_f)$ denote
the number of instances belonging to category $c\in\mathcal C$, where
$\mathcal C$ is the semantic-category vocabulary.

Accordingly, the task-level image $I$ defined in
Section~\ref{sec:task:formal} corresponds to the source frame $I_f$.

\subsubsection{Stage 1 --- Frame Filtering}
\label{sec:dataset:stage1}

Stage~1 identifies frames that support unambiguous, multi-instance
CIDQ queries. A frame is retained if it contains at least
$N_{\min}=2$ instances of a query category $C_q$ and exactly one
instance of a distinct reference category $C_r$. Accordingly, the retained frame set is
\begin{equation}
  \mathcal{F}_1
  =
  \left\{
    f\in\mathcal{F}_0
    \;\middle|\;
    \begin{aligned}
      &\exists\,C_q,C_r\in\mathcal{C},
      \quad C_q\neq C_r,\\[-1pt]
      &n_{C_q}\!\left(\mathcal{B}^{\mathrm{gt}}_f\right)
        \geq N_{\min},
      \quad
      n_{C_r}\!\left(\mathcal{B}^{\mathrm{gt}}_f\right)
        =1
    \end{aligned}
  \right\}.
  \label{eq:frame_eligibility}
\end{equation}

Here, $C_q$ and $C_r$ establish only the structural eligibility of a
frame; Stage~4 subsequently enumerates all valid query--reference
configurations. The same eligibility conditions are re-applied after
Stages~2 and~3, as either stage may remove a required candidate or
reference annotation.

\subsubsection{Stage 2 --- LLM-Based Visibility Verification}
\label{sec:dataset:stage2}

For each frame $f\in\mathcal{F}_1$ and annotation
$B_j\in\mathcal{B}^{\mathrm{gt}}_f$, we project the annotated 3D box
onto the RGB image $I_f$ using the calibrated camera parameters
$(\mathbf{K}^{\mathrm{gt}}_f,
\mathbf{R}^{\mathrm{gt}}_f,
\mathbf{t}^{\mathrm{gt}}_f)$, obtaining the projected 2D box
$b^{\mathrm{proj}}_{f,j}$. The corresponding image crop is evaluated by
GPT-4o~\cite{hurst2024gpt4o} using a fixed zero-shot visibility
prompt.

The model assigns each instance a visibility score
$v_{f,j}\in[0,1]$, reflecting whether it is recognizable and
sufficiently visible in the image. We retain annotations satisfying
$v_{f,j}\geq\theta_v$, where $\theta_v=0.7$, and denote the resulting
visibility-verified annotation set by
$\widetilde{\mathcal{B}}_f$. This filtering removes instances that
are heavily occluded, truncated, or visually ambiguous.

After visibility verification, the structural eligibility conditions
defined in Stage~1 are re-applied to
$\widetilde{\mathcal{B}}_f$. Frames that continue to contain a valid
multi-instance query category and an unambiguous reference category
form the retained set $\mathcal{F}_2$.
\subsubsection{Stage 3 --- Geometric Refinement}
\label{sec:dataset:stage3}

Stage~3 refines the projected annotations using independent 2D
detections, while preserving the original MM-Spatial 3D geometry.
For each $f\in\mathcal{F}_2$, GroundingDINO~\cite{liu2024grounding}
is queried using the categories represented in
$\widetilde{\mathcal{B}}_f$, producing detection boxes
$b^{\mathrm{det}}_{f,p}$ with predicted category labels
$\ell^{\mathrm{det}}_{f,p}$, where $p$ indexes the detections.
We apply class-wise NMS with $\theta_{\mathrm{dup}}=0.85$, followed
by a containment check that removes same-class inner boxes using a
five-pixel tolerance.

The remaining detections are matched one-to-one with the projected
MM-Spatial annotations using class-consistent Hungarian assignment.
A match $(j,p)$ is admissible only if
$\mathrm{label}(B_j)=\ell^{\mathrm{det}}_{f,p}$ and
\begin{equation}
  \mathrm{IoU}\!\left(
    b^{\mathrm{proj}}_{f,j},
    b^{\mathrm{det}}_{f,p}
  \right)
  \geq
  \theta_{\mathrm{match}},
  \qquad
  \theta_{\mathrm{match}}=0.5.
  \label{eq:matching_criterion}
\end{equation}
Annotations with admissible one-to-one matches form
$\widehat{\mathcal{B}}_f$.

GroundingDINO is used only to determine annotation retention; the
semantic labels, gravity-aligned centroids, and metric quantities of
$\widehat{\mathcal{B}}_f$ remain inherited from MM-Spatial.
Re-applying the Stage~1 eligibility conditions yields the final frame
set $\mathcal{F}_3$. Exact refinement details are provided in the
supplementary material.

\subsubsection{Stage 4 --- Spatial Synthesis}
\label{sec:dataset:stage4}

Stage~4 converts the verified annotations into CIDQ records.
For each $f\in\mathcal{F}_3$, an annotation
$B\in\widehat{\mathcal{B}}_f$ defines a task-level object $o(B)$
with category $\mathrm{label}(B)$ and gravity-aligned centroid
$\mathbf{c}(B)$. Each annotation $B_r\in\widehat{\mathcal{B}}_f$ whose category occurs
exactly once serves as an unambiguous reference, while every distinct
category $C_q\neq C_r$ with at least $N_{\min}$ retained instances
defines
\[
\begin{aligned}
o_r &= o(B_r),
\qquad
C_r = \mathrm{label}(B_r),\\
\mathbf{c}_r &= \mathbf{c}(B_r),\\
\mathcal{O}_{C_q}(I_f)
&=
\Bigl\{
o(B_i)\ \Bigm|\ 
\substack{
B_i\in\widehat{\mathcal{B}}_f,\\
\mathrm{label}(B_i)=C_q
}
\Bigr\},\\
q &= (I_f,C_q,o_r).
\end{aligned}
\]

Using $\mathbf{c}_i=\mathbf{c}(B_i)$,
Eqs.~\eqref{eq:cidq_inst_index}--\eqref{eq:cidq_dist} determine the
closest instance $o_{i_q^\star}$ and its ground-truth floor-plane
distance $d_q^\star$, yielding the T1 record $(q,d_q^\star)$.

Each structured configuration $q$ is verbalized as a natural-language
question using one of 100 English templates, with 50 templates per
sub-task. For T2,
\begin{equation}
\begin{aligned}
q_\tau &= (q,\tau),
\qquad
y_{q_\tau}^\star
=
\mathbf{1}\!\left[d_q^\star\leq\tau\right],\\
\tau &\in
\{0.5,1.0,1.5,2.0,3.0\}\,\mathrm{m}.
\end{aligned}
\label{eq:stage4_t2_instantiation}
\end{equation}
This yields the T2 record $(q_\tau,y_{q_\tau}^\star)$.
Thresholds are sampled across distance strata to approximately balance
the binary labels. The final benchmark is
$\mathcal{Q}=\mathcal{Q}_{\mathrm{T1}}\cup\mathcal{Q}_{\mathrm{T2}}$.
All targets inherit the original gravity-aligned MM-Spatial geometry;
predicted depth and camera parameters are not used during synthesis.

\subsection{Benchmark Scale and Comparison}
\label{sec:dataset:stats}

\begin{table}[t]
\centering
\caption{%
  \textbf{Comparison of spatial-reasoning benchmarks.}
  $^\dagger$Ground-truth geometry is used only for offline label
  construction and is unavailable to evaluated models.
}
\label{tab:dataset_comparison}
\resizebox{\columnwidth}{!}{%
\begin{tabular}{@{} l r c c c c c @{}}
\toprule
\textbf{Dataset}
  & \textbf{\#QA}
  & \makecell{\textbf{RGB-}\\\textbf{only}}
  & \makecell{\textbf{Multi-}\\\textbf{inst.}}
  & \makecell{\textbf{Metric}\\\textbf{GT}}
  & \makecell{\textbf{Floor-}\\\textbf{plane GT}}
  & \textbf{Setting} \\
\midrule

VSR~\cite{liu2023visual}
  & 10,972
  & \cmarkbold
  & \ding{55}
  & \ding{55}
  & \ding{55}
  & In./Out. \\

SpatialBench~\cite{chen2024spatialvlm}
  & 100,000
  & \cmarkbold
  & \ding{55}
  & \cmarkbold
  & \ding{55}
  & In./Out. \\

SQA3D~\cite{ma2023sqa3d}
  & 33,400
  & \ding{55}
  & \ding{55}
  & \ding{55}
  & \ding{55}
  & Indoor \\

EmbodiedScan~\cite{wang2024embodied}
  & 1,694,723
  & \ding{55}
  & \ding{55}
  & \cmarkbold
  & \ding{55}
  & Indoor \\

\midrule
\rowcolor{rowgray}
\textbf{\dataset{} (ours)}
  & \textbf{1,064,022}
  & \cmarkbold
  & \cmarkbold
  & \cmarkbold$^\dagger$
  & \cmarkbold$^\dagger$
  & \textbf{Indoor} \\

\bottomrule
\end{tabular}%
}
\end{table}

\dataset{} contains 1,064,022 QA pairs generated from 73,924 frames
across 200 indoor scenes.
The benchmark comprises 659,786 T1 distance-estimation records
(62.0\%) and 404,236 T2 proximity-decision records (38.0\%), spanning
315 object categories.
The T1 floor-plane distances have a mean of 1.52\,m, a median of
1.15\,m, and a 95th percentile of 3.95\,m, covering both near-field
and mid-range indoor configurations.

All main experiments use fixed task-specific evaluation subsets drawn
from this benchmark under the controlled protocol specified in
Section~\ref{sec:exp:setup}.

\section{Experiments}
\label{sec:experiments}

\subsection{Experimental Setup}
\label{sec:exp:setup}

We evaluate \ours{} and all baselines under a common zero-shot
protocol, with no task-specific fine-tuning, ground-truth depth, or
calibrated camera intrinsics available at inference time.
Qwen3-VL-8B is used as the primary backbone, while
Qwen2.5-VL-3B assesses cross-backbone generalisation. All open-source
experiments are conducted on a single NVIDIA RTX 5880 Ada Generation, 48 GB. We
compare pure RGB-based VLMs, closed-source frontier models, and
fine-tuned spatial specialists using identical decoding settings.

We construct fixed common evaluation sets of 5{,}000 T1 queries and
5{,}000 T2 queries sampled from \dataset{}. Both sets cover all 200
source scenes and are stratified by object category, floor-plane
distance, and task-specific factors. For T2, 1{,}000 queries are selected
for each threshold
$\tau\in\{0.5,1.0,1.5,2.0,3.0\}$\,m, with balanced positive and
negative labels.

\noindent\textbf{Black-image sanity filter.}\quad
Each query is evaluated using both the original image and a
same-resolution uniform black image. For an accuracy-style metric, a
query receives credit only when the method returns valid outputs under
both conditions, is correct on the original image, and is incorrect on
the black image. Invalid, missing, or unparsable outputs receive zero
credit. All accuracy-style metrics therefore use the complete fixed
task-specific set of 5{,}000 queries as the denominator. Floor-MAE is
computed only from valid original-image distance predictions.

\begin{table*}[t]
\centering
\caption{%
  \textbf{T1: Closest-Instance Distance Estimation} on a fixed common
evaluation set of 5{,}000 queries covering all 200 source scenes.
  Floor-MAE is reported in metres ($\downarrow$); Acc and Unc-Acc are
  percentages ($\uparrow$). Accuracy-style metrics use the black-image
  sanity filter. Methods with the \ours{} pipeline use geometry-derived
  BEV uncertainty, whereas the remaining methods use prompt-elicited
  uncertainty. Bold indicates the best overall result.
}
\label{tab:t1_distance}
\resizebox{\textwidth}{!}{%
\begin{tabular}{@{} l c c c c c c c @{}}
\toprule
\multirow{2}{*}{\textbf{Method}}
  & \multirow{2}{*}{\textbf{Train-Free}}
  & \multicolumn{3}{c}{\textbf{Deterministic Metrics}}
  & \multicolumn{3}{c}{\textbf{Uncertainty-Aware Metrics}} \\
\cmidrule(lr){3-5}\cmidrule(lr){6-8}
  &
  & \textbf{Floor-MAE$\downarrow$}
  & \textbf{Acc@10\%$\uparrow$}
  & \textbf{Acc@0.2m$\uparrow$}
  & \textbf{Unc-Acc@10\%$\uparrow$}
  & \textbf{Unc-Acc@0.2m$\uparrow$}
  & \textbf{Unc-Acc@0.3m$\uparrow$} \\
\midrule
\multicolumn{8}{@{}l}{\textit{Pure VLM Baselines (zero-shot, no geometric pipeline)}} \\[2pt]
Qwen2.5-VL-3B~\cite{bai2025qwen25vl}
  & \cmarkbold & 0.5354 & 9.50 & 26.50
  & 18.24 & 27.54 & 33.42 \\
Qwen2.5-VL-7B~\cite{bai2025qwen25vl}
  & \cmarkbold & 0.498 & 1.50 & 15.50
  & 13.00 & 21.00 & 24.00 \\
Qwen3-VL-8B
  & \cmarkbold & 0.432 & 7.50 & 24.50
  & 15.50 & 26.50 & 36.00 \\
\midrule
\multicolumn{8}{@{}l}{\textit{Closed-Source Models }} \\[2pt]
GPT-4o-mini~\cite{openai2024gpt4omini}
  & \cmarkbold & 0.402 & 11.40 & 27.86
  & 11.40 & 27.86 & 34.18 \\
GPT-4o~\cite{hurst2024gpt4o}
  & \cmarkbold & 0.481 & 1.66 & 6.04
  & 6.60 & 13.20 & 19.24 \\
Gemini-2.5-Flash~\cite{comanici2025gemini25}
  & \cmarkbold & 0.495 & 2.98 & 11.32
  & 11.32 & 21.44 & 25.00 \\
Gemini-2.5-Pro~\cite{comanici2025gemini25}
  & \cmarkbold & 0.448 & 6.32 & 13.68
  & 10.54 & 21.06 & 26.32 \\
\midrule
\multicolumn{8}{@{}l}{\textit{Fine-tuned / Specialized Models (spatial reasoning)}} \\[2pt]
SpaceThinker-3B~\cite{remyx2025spacethinker}
  & \xmark & 1.254 & 14.00 & 39.50
  & 33.00 & 46.00 & 57.50 \\
SpaceOm~\cite{remyx2025spaceom}
  & \xmark & 1.049 & 14.50 & 28.50
  & 21.50 & 35.00 & 47.50 \\
Spatial-SSRL-7B~\cite{liu2025spatial}
  & \xmark & 0.454 & 4.50 & 17.00
  & 4.50 & 17.00 & 32.00 \\
Spatial-SSRL-Qwen3VL-4B~\cite{liu2025spatial}
  & \xmark & 0.357 & 10.82 & 42.00
  & 12.50 & 43.00 & 54.00 \\
\midrule
\multicolumn{8}{@{}l}{\textit{\textsc{SpatialQuery} (Ours, zero-shot, train-free geometric pipeline)}} \\[2pt]
\textsc{SpatialQuery} w/ Qwen2.5-VL-3B
  & \cmarkbold
  & 3.643 & 10.82 & 31.50
  & 19.50 & 38.50 & 53.00 \\
\rowcolor{gray!20}
\textsc{SpatialQuery} w/ Qwen3-VL-8B
  & \cmarkbold
  & \textbf{0.259} & \textbf{34.00} & \textbf{58.50}
  & \textbf{70.50} & \textbf{87.50} & \textbf{90.50} \\
\bottomrule
\end{tabular}%
}
\end{table*}

\begin{table}[t]
\centering
\caption{%
  \textbf{T2: Proximity Decision Classification} on the fixed
  5{,}000-query subset
  $\mathcal{Q}_{\mathrm{eval}}^{\mathrm{T2}}$.
  Accuracy is reported after the black-image sanity filter.
  Bold indicates the best result within each method group.
}
\label{tab:t2_proximity}
\begin{tabular}{@{} l c @{}}
\toprule
\textbf{Method} & \textbf{Accuracy (\%)} \\
\midrule
\multicolumn{2}{@{}l}{\textit{Pure VLM Baselines}} \\[2pt]
Qwen2.5-VL-3B  & \textbf{64.90} \\
Qwen2.5-VL-7B  & 41.62 \\
Qwen3-VL-8B    & 59.90 \\
\midrule
\multicolumn{2}{@{}l}{\textit{Closed-Source Models }} \\[2pt]
GPT-4o-mini~\cite{openai2024gpt4omini}  & 54.46 \\
GPT-4o~\cite{hurst2024gpt4o}  & 60.26 \\
Gemini-2.5-Flash~\cite{comanici2025gemini25}  & 52.52 \\
Gemini-2.5-Pro~\cite{comanici2025gemini25}  & \textbf{65.24} \\
\midrule
\multicolumn{2}{@{}l}{\textit{Fine-tuned / Specialized Models}} \\[2pt]
SpaceThinker-3B              & 46.68 \\
SpaceOm                      & 66.68 \\
Spatial-SSRL-7B              & 53.38 \\
Spatial-SSRL-Qwen3VL-4B      & \textbf{76.04} \\
\midrule
\multicolumn{2}{@{}l}{\textit{\textsc{SpatialQuery} (Ours, train-free)}} \\[2pt]
\textsc{SpatialQuery} w/ Qwen2.5-VL-3B  & 68.52 \\
\rowcolor{gray!20}
\textsc{SpatialQuery} w/ Qwen3-VL-8B    & \textbf{84.18} \\
\bottomrule
\end{tabular}
\end{table}

\subsection{Research Questions and Main Results}
\label{sec:exp:main_results}

\paragraph{RQ1: Can training-free geometric grounding outperform
specialized spatial models?}
As shown in Table~\ref{tab:t1_distance}, \ours{} with Qwen3-VL-8B
achieves the lowest Floor-MAE among valid predictions, reducing the
error from 0.357\,m for the strongest specialized baseline to
0.259\,m, a relative reduction of 27.5\%. It also improves the
sanity-filtered Acc@0.2\,m from 42.00\% to 58.50\%, corresponding to
a gain of 16.50 percentage points, despite requiring neither
task-specific fine-tuning nor architectural modification.

These results suggest that spatial specialization alone does not
necessarily transfer to CIDQ. In contrast to conventional fixed-pair
reasoning, CIDQ requires a model to recover a variable-size candidate
set, compare all candidate--reference relations, and select the
minimum-distance instance. Explicitly reconstructing these instances
in a common metric floor-plane frame therefore provides a task-aligned
inductive structure that is difficult to recover through implicit
RGB-only reasoning alone.

\paragraph{RQ2: Does geometry-derived uncertainty identify unreliable
predictions and improve error coverage?}
Conditioning the acceptance tolerance on the geometry-derived
uncertainty $\sigma_q$ increases Acc@0.2\,m from 58.50\% to an
Unc-Acc@0.2\,m of 87.50\%, while Unc-Acc@0.3\,m reaches 90.50\%.
The additional 3.00-point gain obtained by increasing the base
tolerance from 0.2\,m to 0.3\,m indicates that most uncertainty-aware
coverage is already obtained within the tighter 0.2\,m setting.

Importantly, increased coverage alone does not establish that
$\sigma_q$ is an informative reliability signal, since any positive
uncertainty margin enlarges the acceptance interval. The
uncertainty-stratified analysis in the supplementary material provides
more direct evidence: the high-dispersion subset yields a Floor-MAE of
0.349\,m, compared with 0.218\,m for the low-dispersion subset, an
increase of approximately 60\%. Thus, $\sigma_q$ is informative for
ranking geometrically difficult queries, although it should be
interpreted as a reliability proxy rather than a calibrated
probabilistic confidence interval.

\paragraph{RQ3: How sensitive is the framework to backbone choice?}
The results reveal substantial backbone sensitivity. Replacing
Qwen3-VL-8B with Qwen2.5-VL-3B in the full pipeline increases
Floor-MAE from 0.259\,m to 3.643\,m and reduces Acc@0.2\,m from
58.50\% to 31.50\%. Nevertheless, relative to the corresponding pure
Qwen2.5-VL-3B baseline, the geometric pipeline improves Acc@0.2\,m
from 26.50\% to 31.50\%, Unc-Acc@0.3\,m from 33.42\% to 53.00\%,
and T2 accuracy from 64.90\% to 68.52\%.

The simultaneous improvement in tolerance-based metrics and
degradation in Floor-MAE indicates a non-monotonic effect: the smaller
backbone benefits from the structured geometric representation on many
queries but occasionally produces severe numerical outliers that
dominate the mean absolute error. The framework is therefore portable
across the two evaluated backbones, but not backbone-invariant;
reliable continuous metric generation still requires sufficient
instruction-following and numerical reasoning capacity.

\paragraph{RQ4: Does recovered metric geometry support proximity
decisions?}
As reported in Table~\ref{tab:t2_proximity}, \ours{} with Qwen3-VL-8B
achieves a T2 accuracy of 84.18\%. This represents a 24.28-point gain
over the corresponding pure Qwen3-VL-8B baseline, and exceeds the
strongest fine-tuned specialist and closed-source model by 8.14 and
18.94 percentage points, respectively. The Qwen2.5-VL-3B pipeline
also improves over its pure-backbone baseline, from 64.90\% to
68.52\%.

Because T2 applies a decision threshold directly to the same
closest-instance distance estimate used in T1, these gains show that
the reconstructed metric geometry transfers to threshold-based
reasoning rather than benefiting distance regression alone. Binary
decisions are insensitive to distance errors that do not cross the
threshold $\tau$, whereas even a small error near the decision
boundary can reverse the label. This explains why method rankings
under T2 accuracy need not exactly follow those under T1 Floor-MAE.

\newcommand{\sd}[1]{{\scriptsize$\,{\pm}#1$}}
\newcommand{\sig}{$^{*}$}

\section{Ablation Studies}
\label{sec:ablation}

Having established the overall effectiveness of \ours{}, we next
isolate how robust geometric refinement, BEV Scene Cubifying, and
\uacot{} contribute to the observed gains. Table~\ref{tab:ablation}
reports an incremental ablation using Qwen3-VL-8B under the same
zero-shot protocol.

\begin{table*}[t]
\centering
\caption{%
  Component ablation on the same fixed common set of 5{,}000 T1 queries from
\dataset{}
  (Qwen3-VL-8B backbone, zero-shot).
  \uacot{} is enabled in all rows R1--R3; R0 is the pure-VLM
  reference with no geometric pipeline and no \uacot{}. Accuracy-style metrics follow the same sanity-filtered evaluation
protocol as the main experiments.
  Floor-MAE in metres~($\downarrow$);
  Acc and Unc-Acc in \%~($\uparrow$). \textbf{Bold}: best per column.
  TC+MR\,=\,Tilt-Corrected MAD-RANSAC;
  BEV\,=\,BEV Scene Cubifying;
  UA\,=\,\uacot{} Prompting.%
}
\label{tab:ablation}
\renewcommand{\arraystretch}{1.20}
\resizebox{\textwidth}{!}{%
\begin{tabular}{@{} r @{\hspace{6pt}} l @{\hspace{8pt}}
                c c c @{\hspace{6pt}}
                r r r @{\hspace{6pt}}
                r r @{}}
\toprule
& &
  \multicolumn{3}{c}{\textsc{Component}} &
  \multicolumn{3}{c}{\textsc{Deterministic (T1)}} &
  \multicolumn{2}{c}{\textsc{Uncertainty-aware (T1)}} \\
\cmidrule(lr){3-5}
\cmidrule(lr){6-8}
\cmidrule(l){9-10}
& \textbf{Variant} &
  \textbf{TC+MR} & \textbf{BEV} & \textbf{UA} &
  \makecell{\textbf{Floor-MAE}$\downarrow$\\\footnotesize(m)} &
  \makecell{\textbf{Acc@10\%}$\uparrow$\\\footnotesize(\%)} &
  \makecell{\textbf{Acc@0.2m}$\uparrow$\\\footnotesize(\%)} &
  \makecell{\textbf{Unc@0.2m}$\uparrow$\\\footnotesize(\%)} &
  \makecell{\textbf{Unc@0.3m}$\uparrow$\\\footnotesize(\%)} \\
\midrule
\multicolumn{10}{@{}l}{%
  \textit{Reference: no geometric pipeline, no \uacot{}}%
} \\[2pt]
R0 & Pure VLM (Qwen3-VL-8B, raw prompt)
   & \xmark & \xmark & \xmark
   & $0.432$ & $7.50$ & $24.50$
   & $26.50$ & $36.00$ \\
\midrule
\multicolumn{10}{@{}l}{%
  \textit{\uacot{} enabled in all rows below; geometric components
  ablated incrementally}%
} \\[2pt]
R1 & \uacot{} only (no TC+MR, no BEV)
   & \xmark & \xmark & \cmark
   & $16.341$ & $16.00$ & $49.50$
   & $68.00$ & $73.00$ \\
R2 & \hspace{4pt}$+$ Tilt-Corrected MAD-RANSAC
   & \cmark & \xmark & \cmark
   & $0.263$ & $33.00$ & $54.00$
   & $82.50$ & $88.00$ \\
R3 & \hspace{4pt}$+$ BEV Scene Cubifying \quad \textbf{(Full model)}
   & \cmark & \cmark & \cmark
   & $\mathbf{0.259}$ & $\mathbf{34.00}$ & $\mathbf{58.50}$
   & $\mathbf{87.50}$ & $\mathbf{90.50}$ \\
\midrule
\multicolumn{10}{@{}l}{%
  \textit{Cross-backbone generalisation --- Full model (R3)}%
} \\[2pt]
& Qwen2.5-VL-3B~\cite{bai2025qwen25vl}
   & \cmark & \cmark & \cmark
   & $3.643$ & $10.82$ & $31.50$
   & $38.50$ & $53.00$ \\
\rowcolor{gray!20}
& Qwen3-VL-8B \textit{(primary)}
   & \cmark & \cmark & \cmark
   & $\mathbf{0.259}$ & $\mathbf{34.00}$ & $\mathbf{58.50}$
   & $\mathbf{87.50}$ & $\mathbf{90.50}$ \\
\bottomrule
\end{tabular}%
}
\end{table*} 

\paragraph{AQ1: Geometric grounding enables reliable \uacot{} reasoning}

The R0\,$\to$\,R1 transition reveals a non-monotonic effect:
\uacot{} alone increases Floor-MAE from 0.432\,m to 16.341\,m
while improving Acc@0.2\,m from 24.50\% to 49.50\%.
This divergence reflects the different sensitivities of the two
metrics: Floor-MAE is dominated by large numerical outliers, whereas
Acc@$\delta$ only measures whether predictions fall within a fixed
tolerance. Without reliable geometric evidence, structured reasoning
may reinforce erroneous distance estimates and produce rare but
severe failures~\cite{wei2022chain}.

TC+MR resolves this failure by refining the underlying geometric
support. MAD-RANSAC selects the dominant depth mode and removes
boundary and background outliers
(Eqs.~\eqref{eq:mad_ransac_consensus}--\eqref{eq:inlier_point_cloud}),
reducing Floor-MAE from 16.341\,m to 0.263\,m. The retained support
also provides the floor-plane dispersion
$\sigma_i^{\mathrm{BEV}}$ in Eq.~\eqref{eq:bev_uncertainty}, which is
propagated through \uacot{} as a geometry-derived reliability signal.
These results indicate that structured reasoning is effective only
when grounded in sufficiently stable metric geometry.

\paragraph{AQ2: TC+MR and BEV address complementary bottlenecks}

The R2\,$\to$\,R3 transition isolates the contribution of BEV Scene
Cubifying after depth-domain errors have been corrected by TC+MR.
By representing instances as equal-size, category-coded blocks on a
canonical top-down canvas, the BEV suppresses texture, perspective,
and object-scale variation while preserving the relative layout
required for candidate comparison. This abstraction improves
Acc@0.2\,m by 4.5 percentage points and Unc-Acc@0.3\,m by
2.5 points. Relative to the pure-VLM baseline R0, the full model
reduces Floor-MAE by 40.0\% and improves Unc-Acc@0.3\,m by
54.5 points.

The cross-backbone results further show that performance depends on
reasoning capacity. With Qwen2.5-VL-3B, the full pipeline yields a
higher Floor-MAE of 3.643\,m, suggesting less stable continuous
numerical reasoning at the smaller model scale. Nevertheless, it
improves Acc@0.2\,m from 26.50\% to 31.50\% and
Unc-Acc@0.3\,m from 33.42\% to 53.00\% over the corresponding pure
backbone. Thus, the geometric representation remains beneficial,
although accurate metric generation requires sufficient
instruction-following capacity.

\section{Discussion and Limitations}
\label{sec:discussion}
\label{sec:limitations}

The results suggest two broader implications. First, reliable metric
reasoning requires both stable geometric evidence and a task-aligned
representation: structured reasoning cannot correct globally inaccurate
geometry, while the canonical BEV facilitates variable-cardinality
candidate comparison once the reconstructed scene is sufficiently
reliable. Second, spatial fine-tuning on fixed object pairs does not
necessarily transfer to CIDQ, which additionally requires candidate-set
aggregation and closest-instance selection.

\noindent\textbf{Limitations.}\quad
\ours{} depends on monocular metric depth and therefore remains
sensitive to low-texture regions, reflective or transparent surfaces,
severe occlusion, and globally inaccurate depth predictions.
MAD-RANSAC can suppress local outliers, and
$\sigma_i^{\mathrm{BEV}}$ can expose dispersed geometry, but neither
can recover information absent from the underlying depth estimate.

The framework also relies on the VLM to interpret structured geometry
and perform stable numerical comparison, making performance sensitive
to backbone capacity. Moreover, the current formulation is restricted
to static, single-floor indoor scenes observed from a single RGB image.
Future work may explore deterministic geometric reasoning modules,
selective model invocation, temporal or multi-view observations, and
explicit occlusion reasoning.

\section{Conclusion}
\label{sec:conclusion}
We presented \ours{}, a training-free framework for the
Closest-Instance Distance Query (CIDQ)---a multi-instance spatial
reasoning task unaddressed by prior benchmarks and VLMs.
The central insight is that metric spatial reasoning requires two
complementary forms of grounding: a viewpoint-agnostic BEV
abstraction (\emph{Scene Cubifying}) that redirects the VLM's
attention from texture noise to floor-plane geometry, and an
uncertainty-aware reasoning chain (\uacot{}) that prevents
overconfident metric decisions in geometrically ambiguous regions.
Neither component alone suffices---geometric stabilisation is a
prerequisite for structured reasoning, and structured reasoning
amplifies the gains of geometric grounding.
On the accompanying million-scale benchmark \dataset{},
\ours{} with a Qwen3-VL-8B backbone achieves a Floor-MAE of
\textbf{0.259\,m} and Unc-Acc@0.3\,m of \textbf{90.5\%} under a
strict zero-shot protocol, outperforming both fine-tuned spatial
specialists and closed-source frontier models without any
task-specific training.
We hope \ours{} and \dataset{} provide a foundation for
geometry-aware, uncertainty-aware spatial reasoning in
embodied and assistive AI systems.

\bibliographystyle{IEEEtran}
\bibliography{ref}
\end{document}